\documentclass[conference]{IEEEtran}
\usepackage{booktabs}
\usepackage{comment} 
\usepackage{cite}
\usepackage{amsmath,amssymb,amsfonts}
\usepackage{algorithmic}
\usepackage{graphicx}
\usepackage{textcomp}
\usepackage{xcolor}
\usepackage[hidelinks]{hyperref} 
\def\BibTeX{{\rm B\kern-.05em{\sc i\kern-.025em b}\kern-.08em
    T\kern-.1667em\lower.7ex\hbox{E}\kern-.125emX}}
\begin{document}
% \showthe\linewidth
% \showthe\floatsep
% \showthe\intextsep
% \showthe\textfloatsep
% \showthe\dbltextfloatsep
% \showthe\dblfloatsep
% \showthe\abovecaptionskip
% \showthe\belowcaptionskip
% \setlength{\abovecaptionskip}{1.5pt} %peng
\setlength{\intextsep}{0.5pt} %peng, Space above and below in-text floats (figures/tables placed within text).
\setlength{\floatsep}{0pt} %peng, Space between two floats that appear together (e.g., two figures in a row).
\setlength{\textfloatsep}{0.5pt} %peng, Space between floats that appear at the top or bottom of a page and the surrounding text.

\title{Spatial Language Likelihood Grounding Network for Bayesian Fusion of Human-Robot Observations\\
}

\author{
    \IEEEauthorblockN{
    Supawich Sitdhipol\textsuperscript{1}\textsuperscript{,}\textsuperscript{2}\IEEEauthorrefmark{2}, 
    Waritwong Sukprasongdee\textsuperscript{1}\textsuperscript{,}\textsuperscript{3}\IEEEauthorrefmark{2}, 
    Ekapol Chuangsuwanich\textsuperscript{2}, 
    Rina Tse\textsuperscript{1}\textsuperscript{,}\textsuperscript{3}\IEEEauthorrefmark{1}
    }
}

\maketitle

\begingroup\renewcommand\thefootnote{}
\footnotetext{\IEEEauthorrefmark{2}Equal contribution. \IEEEauthorrefmark{1}Corresponding author: Rina.T@chula.ac.th}
\footnotetext{\{\textsuperscript{1}Autonomous Systems Lab, \textsuperscript{2}Dept. of Computer Engineering, \textsuperscript{3}Dept. of Mechanical Engineering\}, Faculty of Engineering, Chulalongkorn University, Thailand}
\footnotetext{Supplementary video available at \href{https://cu-asl.github.io/fp-lgn}{https://cu-asl.github.io/fp-lgn}}
%\href{https://github.com/cu-asl}{https://github.com/cu-asl}}
\endgroup
\begin{abstract} % At most 3000 characters are allowed.
Fusing information from human observations can help robots overcome sensing limitations in collaborative tasks. 
However, an uncertainty-aware fusion framework requires a grounded likelihood representing the uncertainty of human inputs.
This paper presents a Feature Pyramid Likelihood Grounding Network (FP-LGN) that grounds spatial language by learning relevant map image features and their relationships with spatial relation semantics.
The model is trained as a probability estimator to capture aleatoric uncertainty in human language using three-stage curriculum learning.
Results showed that FP-LGN matched expert-designed rules in mean Negative Log-Likelihood (NLL) and demonstrated greater robustness with lower standard deviation. %than other models. 
\textcolor{black}{Collaborative sensing results demonstrated that the grounded likelihood successfully enabled uncertainty-aware fusion of heterogeneous human language observations and robot sensor measurements, achieving significant improvements in human-robot collaborative task performance.}
\end{abstract}
\vspace{-4 pt}

\vspace{-2 pt}
\section{Introduction}
\vspace{-4 pt}
One key challenge in autonomous robotics is enabling a robot to perform perception tasks as well as subsequent reasoning and decision-making under uncertainty \cite{probRoboticsBook, BarShalom, TseRAL19_short}.
As such, uncertainty-aware perception 
techniques are crucial for improving the robustness of an autonomous robot's performance. 
Furthermore, to make an optimal decision in complex scenarios, a robot must be able to fuse information from multiple, oftentimes heterogeneous sources to update its knowledge.  
Numerous techniques have been developed to optimally fuse information from a diverse set of sensors according to measurement uncertainty in many usage scenarios \cite{fusion_PF_AUV_Tracking_RAL24,fusion_ESKF_CuRobot_localize_RAL24,Wyffels2015-ul}. These techniques are often based on a Bayesian principle \cite{BarShalom,probRoboticsBook,Wakayama2023}, where a robot’s posterior belief is updated according to each measurement's uncertainty, modeled by a measurement likelihood distribution.
Bayesian fusion of information from heterogeneous sources has been found useful in many robotic tasks, such as 
localization 
\cite{EKF_IMU_Odom_2DLiDAR_IROS_2020, Charrow2014-tq}, %\cite{fusion_ESKF_CuRobot_localize_RAL24} 
target tracking \cite{KF_singleCam_MOT_DataAssoc_IROS22,EKFvisualTrackingICRA24,ferrari23fusion}, and mapping \cite{TseTRO2015,TseMFI2012}. 
\textcolor{black}{One key advantage of Bayesian fusion is its recursive formulation, where  each new observation updates the belief incrementally while preserving and propagating uncertainty.}
In traditional approaches, these frameworks have been developed for physical sensor inputs. However, previous developments \cite{Ahmed2013-mm,Wakayama2023} have attempted to integrate natural language observation inputs from humans into the existing Bayesian perception framework.

\vspace{-5pt}

\begin{figure}[t]
\centering
\includegraphics[width=3in]{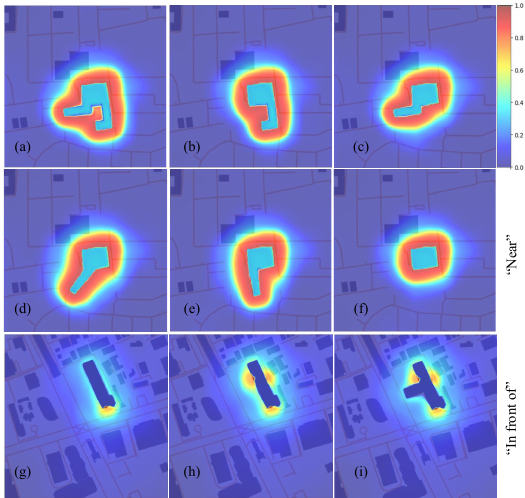}
\vspace{-8pt}
\caption{Learned likelihood distributions of spatial language inputs, given varying landmark features and environmental contexts. This likelihood grounding can be flexibly used in recursive Bayesian fusion frameworks to probabilistically integrate human observations with heterogeneous robot sensor measurements, while explicitly handling input uncertainties. Notice the fully-learned multimodal nature of ``in front of" likelihood distributions for buildings with multiple entrances. A mixture distribution with high-probability regions predicted in front of each entrance and lower probability regions around the building was observed.
This result reflects the model's ability to automatically capture the language input uncertainty due to the ambiguity in human spatial semantics.}
\label{fig:first_spatial_language_and_likelihood}
\end{figure}

\vspace{-1pt}
Human observation inputs are often essential for helping robots overcome sensing constraints. Previous research \cite{Ahmed2013-mm,NisarMMS_ACC08,NisarICRA2010,AhmedAAAI14} have proposed human-robot collaborative sensing paradigms that integrate data from both traditional sensors and human-generated spatial observation statements for robot belief updates. The approach has shown promising results in enhancing decision-making within human-robot collaborative information gathering tasks such as in target search applications. It was found that the combination of human and robot sensor data can reduce search time compared to robots using only sensor data. 

During a human-robot collaborative Bayesian information gathering process \cite{TseTRO2018, TseIROS2015}, a robot recursively grounds a human spatial language input to a likelihood distribution according to the degree of uncertainty corresponding to the statement's semantics. With a grounded likelihood, a natural language input can then be seamlessly integrated into existing Bayesian fusion techniques.
To achieve this goal, research has been conducted on likelihood modeling of spatial language. 
In previous work \cite{Ahmed2013-mm,AhmedAAAI14}, the likelihood of human spatial language input was represented via a Multimodal Softmax (MMS) model learned using maximum likelihood estimation on human-annotated data \cite{NisarExpertSyst2012}.
An efficient recursive Bayesian fusion framework for combining robot sensor measurements with human-generated inputs using a Variational Bayesian Importance Sampling (VBIS) technique was developed. 
Similarly, the work in \cite{BishopLanguage2013, BishopFusion2011, BishopRistic2011} proposed a recursive Bayesian update with spatial language expressions via a random set modeling for spatial language likelihood functions.
A modeling of spatial language likelihood was also proposed in \cite{arulampalam18spatial_natural_language_binary_model} using a Bayesian method in deriving the theoretical posterior Cramer-Rao lower bound to estimate parameters in binary models for ``near'' spatial relationship.

\vspace{-2 pt}%add
However, a major challenge remained since the likelihood distribution reflecting the semantics of a spatial relation depends on the context, e.g., the geometric properties of the landmark used by human in describing the spatial information. 
This contextual information is embedded in the map of the environment. 
Thus, the likelihood grounding of a spatial language input must be trained to extract the relevant features from the environment map and adapt the likelihood distribution of the input spatial language according to the context given.
To illustrate this concept, \figurename~\ref{fig:first_spatial_language_and_likelihood} shows the variations in likelihood grounding of spatial language sentences: (a)-(f) ``The subject is near the building,'' and (g)-(i) ``The subject is in front of the building,'' given a variety of environment maps. 
Two original buildings shown in \figurename~\ref{fig:first_spatial_language_and_likelihood}(a) and (g) were extracted from OpenStreetMap (OSM) \cite{OpenStreetMap}. 
The building in \figurename~\ref{fig:first_spatial_language_and_likelihood}(a) was modified such that its wings were either removed or rotated. 
Similarly, the building in \figurename~\ref{fig:first_spatial_language_and_likelihood}(g) was modified to add additional entrances locations from one to three, as well as an additional protrusion in its shape. 
The likelihood grounding results adapted to the modified contextual map and are shown in \figurename~\ref{fig:first_spatial_language_and_likelihood}(b)-(f) and 
\figurename~\ref{fig:first_spatial_language_and_likelihood}(h)-(i).

\vspace{-1.5 pt}%add
To integrate additional contextual map information into spatial language likelihood modeling, 
the work in \cite{Sweet2016-jm, Ahmed2018-mw} solved data-free and data-sparse likelihood synthesis problems by incorporating the geometric attributes of the known landmarks as constraints in the multimodal softmax parameter estimation. 
Also, a batch fusion update was developed for computational efficiency in \cite{Sweet2016-jm}. The map was augmented with human's sketches, forming representative vertices of the reference landmark used for likelihood grounding in \cite{Burks2023}. 
Subsequently, the work in \cite{Wakayama2023} applied the general spatial language likelihood modeling from \cite{Ahmed2013-mm, Ahmed2018-mw} for recursive Bayesian fusion with probabilistic data association.
However, all of these works relied on predefined expert assumptions about the relationship between the geometry of the log-odds boundaries representing spatial relation semantics and the reference object's geometric features. 
In addition, the work in \cite{Newman2009, Newman2010} proposed a likelihood modeling of spatial language descriptions 
for hidden target's pose, size, and shape estimation. 
The spatial preposition likelihood models were 
written as predefined parametric functions which were then fitted to human empirical data. 
Furthermore, the work in \cite{Zheng2021-zi} introduced a probabilistic spatial language observation modeling for an Object-Oriented POMDP framework. 
A Convolutional Neural Network (CNN) was trained to infer the vector representing the frame of reference, which specified the reference direction upon which a spatial relation %likelihood 
should be computed. 
To determine the likelihood distribution, the likelihood equation was predefined by human experts for each type of spatial relation as a function of the reference landmark's geometric properties based on the concepts from cognitive science research \cite{OKeefe1996, OKeefe2003, Fasola2013-gf}. 

\vspace{0.25 pt}%add
%\vspace{-0.5 pt}%add
Even though the likelihood modeling methods in previous work above enabled the likelihood distribution to adapt to changes in contextual map information, they still relied heavily on predefined assumptions and specifications from human. 
In contrast, this paper proposes a novel spatial language likelihood grounding model, which, to the best of our knowledge, is the first to learn a full adaptation of the likelihood grounding according to the contextual map. 
This is achieved via a Feature Pyramid Likelihood Grounding Network (FP-LGN) which learns the relevant map image features and their relationship with spatial relation semantics.
Unlike previous work, the learning-based approach allows the model to adapt directly to the data, making it more robust to variations and nuances inherent in spatial language semantics. 
The likelihood grounding is useful in enabling the integration of human languages into Bayesian estimation and probabilistic reasoning essential for collaborative human-robot information gathering applications under the presence of uncertainties.

\vspace{0.25 pt}%add
%\vspace{-0.5 pt}%add
The main contributions of this paper are summarized as follows: 
\textbf{(1)} This paper proposes FP-LGN, the first spatial language likelihood grounding model that learns to fully adapt the likelihood to the contextual environment map, trained with three-stage curriculum learning to explicitly model aleatoric uncertainty in human spatial language.
\textbf{(2)} The learned likelihood grounding achieved an information loss performance comparable to the likelihood model written by human experts, showing no statistically significant difference in mean NLL, while exhibiting a greater robustness to variations and nuances in spatial language semantics indicated by the lower standard deviation result.
\textbf{(3)} The likelihood grounding was demonstrated to be successfully used for uncertainty-aware fusion of human language and robot sensor measurements, achieving significant improvements in collaborative sensing task performance. The recursive Bayesian approach allowed interpretable probabilistic reasoning that refined the target posterior over multiple observations, reducing the estimation uncertainty over time.
\vspace{-1pt}

%\vspace{-3.5 pt}%add
\vspace{-3.5 pt}%add
\vspace{-7 pt}
\section{Method}
\vspace{-1 pt}
\vspace{-3.5 pt}%add
%\vspace{-3.5 pt}%add

\subsection{Recursive Bayesian Fusion of Human and Sensor Inputs} \label{subsection:recursive_bayesian_fusion}
\vspace{-4 pt}%add
\vspace{-0.5 pt}%add
This subsection briefly summarizes the general recursive Bayesian updates using robot sensor measurements and human language inputs. Following \cite{Ahmed2013-mm}, let the state of a target of interest at time \(t\) be \(X_t\), a robot sensor measurement be \(Z_t\), and a human spatial language observation be \(S_t\). The recursive Bayesian estimation consists of two steps: prediction and measurement update.
In the prediction step, the target state propagates in time based on its dynamics via the Chapman-Kolmogorov equation, resulting in $p(X_t|Z_{1:t-1},S_{1:t-1})$.
% \begin{IEEEeqnarray}{rCl}
% \IEEEeqnarraymulticol{3}{l}{
% p(X_t|Z_{1:t-1},R_{1:t-1})
% }\nonumber\\* \quad
% & = & \int p(X_t|X_{t-1}) \cdot p(X_{t-1}|Z_{1:t-1},R_{1:t-1}) dX_{t-1}.
% \label{eqn_prediciton_step}
% \end{IEEEeqnarray}
Subsequently, in measurement update steps, each sensor measurement is fused according to its likelihood $p(Z_t|X_t)$:
%test \vspace{-3 pt}
\vspace{-3 pt}%add
\begin{IEEEeqnarray}{rCl}
\IEEEeqnarraymulticol{3}{l}{
p(X_t|Z_{1:t},S_{1:t-1})
}\nonumber\\* \quad
& = &  \frac{p(Z_t|X_t) \cdot p(X_t|Z_{1:t-1},S_{1:t-1})}{\int p(Z_t|X_t) \cdot p(X_t|Z_{1:t-1},S_{1:t-1}) dX_t}.
\label{eqn_robot_measurement_update_step}
\vspace{-3pt}
\end{IEEEeqnarray}
%
%test \vspace{-3pt}
In the same manner, a spatial language observation likelihood \(p(S_t|X_t)\) is used for human measurement update:
%test \vspace{-5pt}
\vspace{-4pt}
\begin{IEEEeqnarray}{rCl}
\IEEEeqnarraymulticol{3}{l}{
p(X_t|Z_{1:t},S_{1:t})
}\nonumber\\* \quad
& = &  \frac{p(S_t|X_t) \cdot p(X_t|Z_{1:t},S_{1:t-1})}{\int p(S_t|X_t) \cdot p(X_t|Z_{1:t},S_{1:t-1}) dX_t}.
\label{eqn_human_measurement_update_step}
\vspace{-2pt}
\end{IEEEeqnarray}
\vspace{-1pt}
Thus, incorporating human spatial language inputs into the Bayesian sensor fusion framework relies on the modeling of spatial language likelihood, which is discussed next.
\vspace{-3 pt}
%smc \vspace{-5 pt}

\vspace{-4pt}
\subsection{Spatial Language Likelihood Grounding and Loss Function}
\label{subsection:spatial_language_likelihood_and_loss_function}
\vspace{-3.5pt}

Following \cite{Zheng2021-zi}, in human-robot communications, 
spatial information about a target of interest can be conveyed using natural language expressions describing the target's spatial relationships with respect to reference landmarks on a map.
Let \(T_i\) denote the $i$th target of interest, \( X_i \) denote its location, and \( \mathcal{M} \) be the map of the environment. 
An input natural language expression may generally consist of \( K \) spatial observations regarding \(T_i\), denoted as \(S_{i,k}\); $k = 1, ..., K$. %\(S_{i,k}\); \(k = 1, ..., K\).
Each \( S_{i,k} \) describes a spatial relation \( R_{i,k} \) of the target with respect to a corresponding reference landmark \( \gamma_{i,k} \).
These observations can be extracted from the input expression via parsing, and represented as tuples \( (T_i, R_{i,k}, \gamma_{i,k}) \). %, where \( R_{i,k} \) denotes the specific spatial relation.
The likelihood of the collective spatial language observation \( S_i = \{S_{i,1}, ..., S_{i,K}\} \) for target \( T_i \), given its location \( X_i \) and the map \( \mathcal{M} \) is factorized as %, can be factorized as %defined as:
%
%test \vspace{-3pt}
%test \vspace{-0.5pt}
%test \vspace{-3pt}
\begin{equation}
\vspace{-3.5pt}
\label{eq:likelihood_def} % Optional: Added a label for referencing
p(S_i \mid X_i, \mathcal{M}) \propto \prod_{k=1}^{K} p(R_{i,k} \mid X_i, \gamma_{i,k}, \mathcal{M}),
%test \vspace{-0.5pt}
%test \vspace{-1pt}
\end{equation}
%\vspace{-3pt}
%
where \( p(R_{i,k} \mid X_i, \gamma_{i,k}, \mathcal{M}) \) represents the likelihood associated with the {uncertainty} of each spatial language observation that must be considered by the robot when incorporating human language inputs into its information fusion and decision-making processes. 
This uncertainty is known as aleatoric uncertainty\cite{pmlr-v162-liu22f}, which, in the context of human linguistics, arises from inherent semantic ambiguity and variability in the interpretation of spatial expressions within the human population.

%To address this aleatoric uncertainty in spatial language grounding,
\textcolor{black}{A Feature Pyramid Likelihood Grounding Network (FP-LGN) is proposed as a probability estimator with the objective of estimating the map-dependent likelihood distribution \smash{$p^* \triangleq p(R_{i,k} \mid\! X_i, \gamma_{i,k}, \mathcal{M})$} capturing the aleatoric uncertainty in human spatial expressions.}
Therefore, the model aims to output a predicted likelihood \( \hat{p} \) that estimates the true distribution \( p^* \). To achieve this, the Kullback-Leibler Divergence (KLD) between the predicted and true distributions $\operatorname{KL}(p^* || \hat{p})$ \cite{TseTRO2018} is minimized, which corresponds to minimizing the expected negative log-likelihood (NLL) over the observed data sampled from $p^*$. 
\textcolor{black}{Thus, the NLL loss was used for FP-LGN training.}

%y \vspace{-5 pt}
\subsection{Likelihood Grounding Network Architecture}
\begin{figure}[t]
\centering
\vspace{-2.5pt}
\includegraphics[width=\linewidth]
{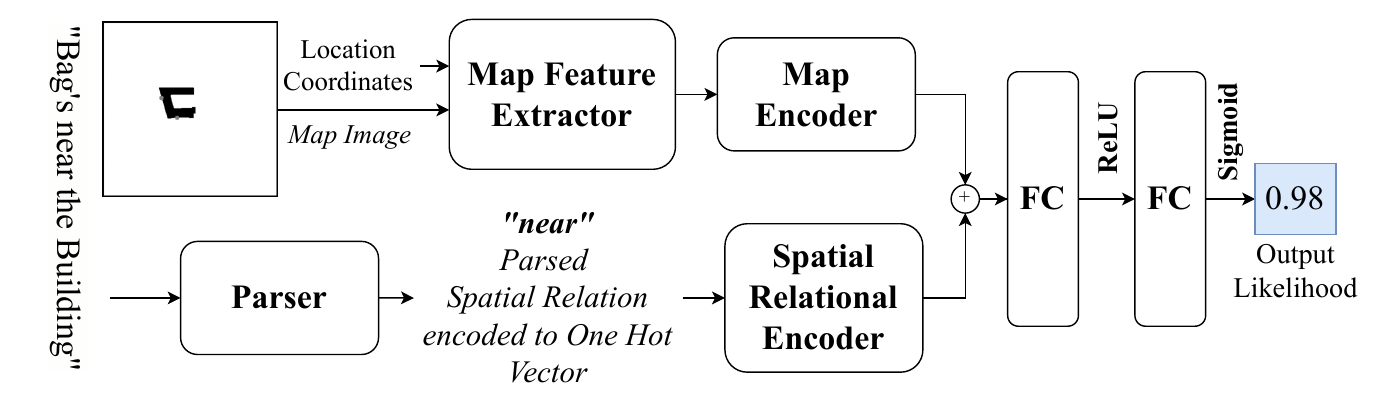}
\vspace{-22pt}
\caption{
The likelihood grounding system, including the parsing module as well as the Feature Pyramid Likelihood Grounding Network (FP-LGN) architecture.}
  \label{fig:model_architecture}
\end{figure}

\vspace{-2.5pt}
\figurename~\ref{fig:model_architecture} provides the overview of the likelihood grounding system including the FP-LGN architecture, which consists of three key components as follows.

\vspace{-0.5pt}
\subsubsection{Map Feature Extractor and Map Encoder}
\begin{figure}[t]
\vspace{-2.5pt}
\centering
\includegraphics[width=\linewidth]{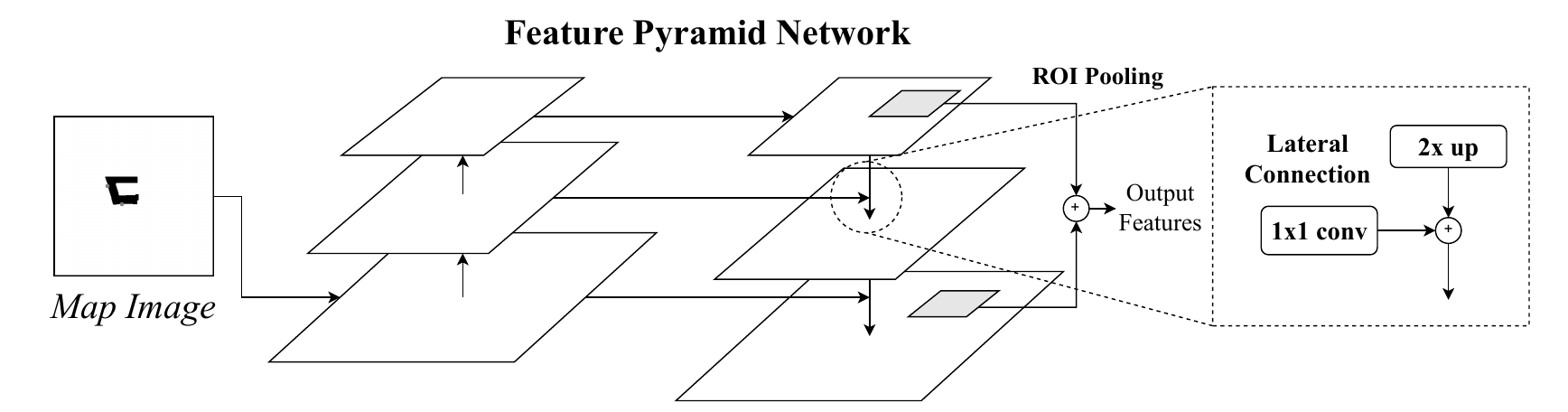}
\vspace{-22pt}
\caption{The FP-LGN map feature extractor which utilizes an FPN to extract features from the map image, the queried location was included through ROI pooling, aiming to capture important features needed to determine the spatial relation around the given location. 
}
\label{fig:feature_extractor}
\end{figure}

Accurate grounding of spatial relations relies on geometric landmark features at multiple levels of details. Thus, maintaining multi-level resolution in the feature extractor becomes crucial. 
To address this, a feature extractor based on the Feature Pyramid Network (FPN) \cite{Lin2016FeaturePN} is proposed. FPN is capable of utilizing information from different resolutions, as illustrated in \figurename~\ref{fig:feature_extractor}. This allows the model to capture both fine and coarse details necessary for various spatial relations. Next, Region of Interest (ROI) Pooling was performed on the feature map layers \cite{ren2016faster}, followed by average pooling on the resulting output. These operations help reduce the feature dimensionality while retaining important spatial information. The pooled features were then passed through dense layers in the encoder, with the final output used for concatenation with other inputs. This design works effectively because it allows the model to adapt to spatial relations of varying complexity, ensuring that the resolution-based details are preserved and utilized.

%\vspace{1.5pt}
\subsubsection{Spatial Relational Encoder}
Following the natural language parsing described in Sec.~\ref{training_parsing}, the extracted spatial relation (\( R_{i,k} \)) is encoded into a one-hot vector representation. This vector is then input into a spatial relation encoder composed of fully connected layers. The output from this encoder serves as a feature for the subsequent stages of the model. This simple yet effective approach allows the model to easily scale and adapt to different variations of spatial relations.

\vspace{-0.5pt}
\vspace{-1pt}
\subsubsection{Text and Map Embedding Interaction}
After extracting feature embeddings from the text and map components, the model combines these embeddings to predict the likelihood output of the spatial relation. This output corresponds to the model's estimation of the probability that the semantics of a particular spatial relationship holds with respect to the contextual reference information in the scene. The two embeddings were concatenated to create a dense representation of the interaction between language and contextual map information before applying it to a sigmoid function at the final layer.
The rationale behind this approach is that multiple spatial relations may have overlapping semantic coverages. For example, a region on a map might be simultaneously described as being near as well as in front of a reference landmark. In this design, the text embedding remains independent from any specific relation, allowing a single location to represent multiple relations simultaneously.

%y \vspace{-2pt}
%test \vspace{-2pt}
\vspace{-5pt}
%smc\vspace{-5pt}
\subsection{Training Methodology and Spatial Relation Parsing}
\label{training_parsing}

\vspace{-2pt}
%\vspace{-2pt}
Since the spatial language likelihood grounding problem differs from traditional classification tasks which do not focus on modeling the aleatoric uncertainty explicitly, our approach aims to train a model that yields a distribution quantifying the probability of a spatial relation given each map location, while explicitly representing this uncertainty as a predicted likelihood~\cite{pmlr-v162-liu22f}.
Learning such distributions requires the model to capture complex contextual information, such as 
the geometric shapes of reference landmarks, while handling the ambiguity of natural language. 
These challenges are compounded by the 
sparsity of human-labeled data, which can lead to unstable or suboptimal results without a proper training strategy.
To address this, a three-stage curriculum-based training strategy that incrementally increases the degree of uncertainty and task complexity is proposed as follows.

%\vspace{-0.5pt}
In the first stage, the model is pretrained using the same strategy as in a regular classification task, i.e., mainly focusing on learning the decision boundary necessary for performing a point estimation of spatial relation class label, rather than estimating the full human semantics likelihood distribution.
This is done by pretraining the model on traditional classification synthetic data given a variety of map images,
allowing the model to focus on learning the relevant feature extractors for various geometric structures in different types of contextual map information.
The second stage of pretraining uses a data synthesis model modified for uncertainty estimator learning by allowing uncertainty in the synthesized labels. This stage uses repeated sampling at each 
training input value to ensure adequate data density for capturing the aleatoric uncertainty representing the ambiguity in human semantics. 
The final stage learns fine-tuned likelihood from synthetic to real aleatoric uncertainty, by fine-tuning the model on real data collected from humans.
%
%y \vspace{-2pt}
%y \vspace{-2pt}
%\vspace{-1pt}
This three-stage curriculum learning approach of gradually increasing complexity and label uncertainty is found to be crucial in allowing a likelihood estimator model to converge to the optimal parameters. 
The Adam optimizer was employed with an initial learning rate of $5 \times 10^{-5}$. A StepLR scheduler was used, with a step size of 10 and a decay factor of 0.6. Early stopping was applied with a patience of 20 epochs.

%y \vspace{-2pt}
\vspace{-1pt}
%y \vspace{-3pt}
To extract spatial observation tuples \((T_i, R_{i,k}, \gamma_{i,k})\) from human-provided spatial language, techniques similar to parse trees \cite{Zheng2021-zi}, neural sequence models \cite{blukis2019learning}, and probabilistic graphical models \cite{tellex2011understanding} could be used. Recent studies \cite{brown2020language, zakharov2019few, li2023revisiting} demonstrate the effectiveness of Large Language Models (LLMs) for zero-shot tasks.  Building on this, LLaMA 2 7B \cite{touvron2023llama} is employed here to extract spatial relations \( R_{i,k} \), landmarks \( \gamma_{i,k} \), and targets \( T_i \) from natural language inputs. The parsing maps input sentences to a predefined dictionary of targets, relations, and landmarks. For example, the observation 
``The robot is in front of building 1, and a bicycle is near building 2'' yields: 
\(\{\mathord{T_1\!:\ }\text{``robot"},\ \mathord{R_{1,1}\!:\ }\text{``in front of"},\ \mathord{\gamma_{1,1}\!:\ }\text{``Building 1"}\}\), and
\(\{\mathord{T_2\!:\ }\text{``bicycle"},\ \mathord{R_{2,1}\!:\ }\text{``near"},\ \mathord{\gamma_{2,1}\!:\ }\text{``Building 2"}\}\).
As multiple parser options are available, the focus of this paper is on the key unaddressed problem of the physical grounding of a spatial relation to a proper likelihood function \smash{$p(R_{i,k} \mid X_i, \gamma_{i,k}, \mathcal{M})$}.

\begin{figure}[t]
\centering
\includegraphics[width=\linewidth]
{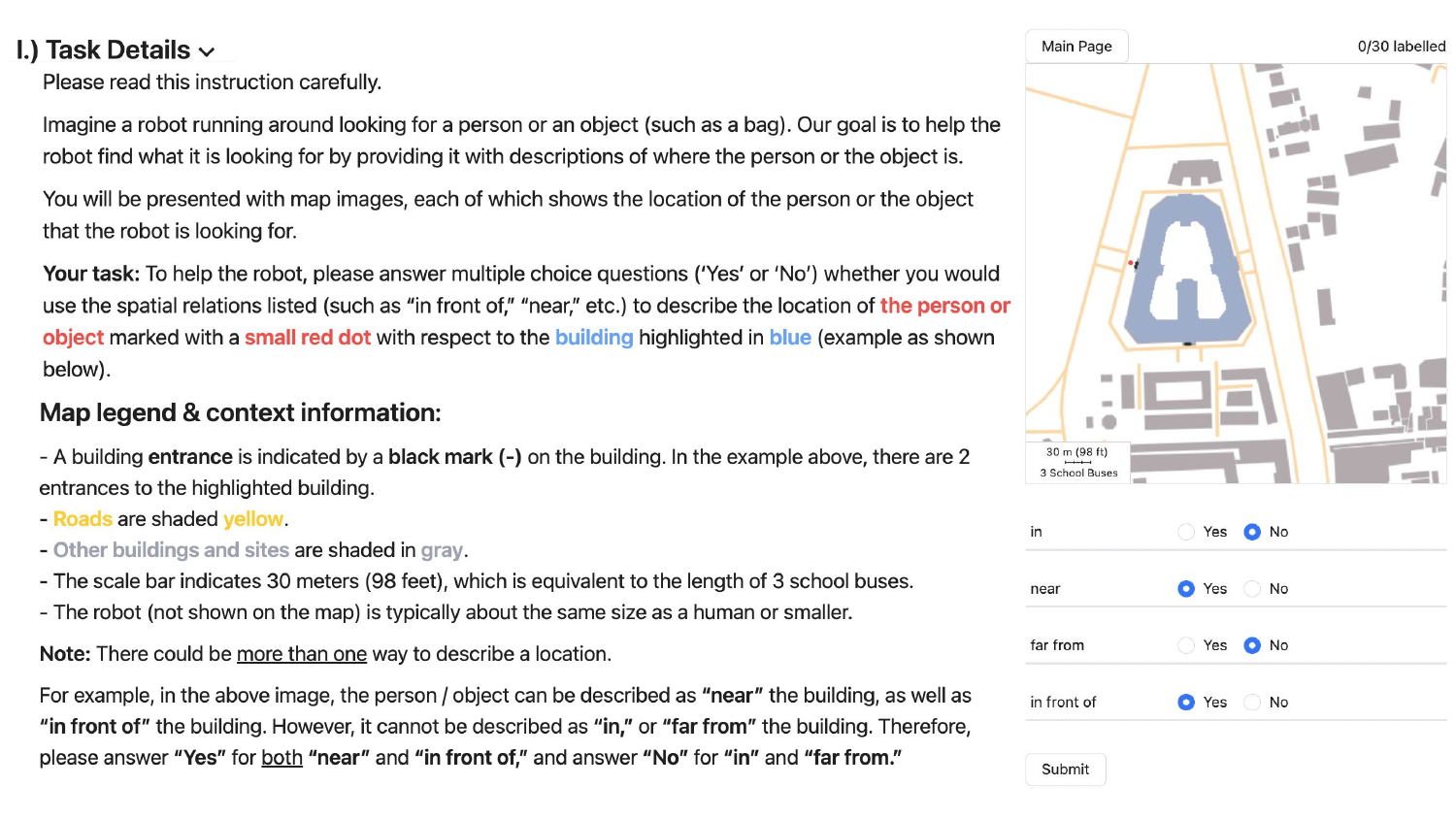}
%test \vspace{-1pt}
\vspace{-25pt}
\caption{The instruction with an example task interface shown to the workers. Each worker could select one or more spatial relations they believed describe the location indicated on the interface.}
\label{fig:data_collection_interface}
\end{figure}
\vspace{-3 pt}
\section{Evaluation}
%\vspace{-1 pt}
\vspace{-2.5 pt}
\subsection{Likelihood Grounding Information Loss}
\label{likelihood_evaluation}
\vspace{-2.5pt}
This evaluation assesses the quality of the likelihood distribution $\hat{p}$ produced by {FP-LGN} in minimizing the information loss when using $\hat{p}$ to represent the underlying distribution of human semantics $p^*$ as quantified by the KLD, $\operatorname{KL}(p^* || \hat{p})$. 
As discussed previously, since KLD minimization corresponds to the maximization of the expected log-likelihood, 
the mean negative log-likelihood (NLL) 
of the unseen human-generated evaluation dataset 
is compared against the benchmark model defined by human experts in previous work.  
An ablation study was also conducted using the same evaluation metric. 

\vspace{-2pt}
First, a spatial language grounding dataset was collected using the Prolific crowdsourcing platform \cite{prolific_2025}. A total of 35 map regions with diverse environments from Bangkok, Thailand, and Washington, DC, USA, were sourced from OSM \cite{OpenStreetMap}. Each region included elements such as buildings, entrances, and streets.
During each task on the Prolific platform, participants were presented with 30 maps, one at a time, to ensure focus. Each map contained a queried location relative to a reference landmark and surrounding context, such as buildings and roads.
Participants assessed whether different spatial relations displayed 
appropriately described the location by answering simple ``Yes'' or ``No'' questions.  
Ten commonly used spatial relations from \cite{Zheng2021-zi}, such as ``at,'' ``next to,'' ``in front of,'' and ``by,'' were used in the evaluation and displayed to the workers at random. 
{A total of 56 crowdworkers participated in the data collection process.}
A screen capture of the labeling interface used for this process is shown in Fig.~\ref{fig:data_collection_interface}.
To maintain data quality while allowing for natural uncertainty in human labeling, two rejection mechanisms were implemented. The first was a worker-centric review, excluding data only if a worker consistently provided inaccurate labels, indicating misunderstanding or inattention (e.g., repeatedly marking points within a building as being ``far from" it). The second operated at the individual data point level, removing only labels that were clearly misassigned (e.g., labeling a location far outside a building as being ``within" it). No majority voting was employed. These measures ensured data reliability while acknowledging inherent semantics variability.
The dataset was split into training and test sets.
%\newline
For each of the 10 spatial relations, a dataset from 21 map regions (2,404 locations) were used for training, while the remaining 14 regions \textcolor{black}{(2,782 locations)} formed the test set. Data augmentation, including random flips and rotations, was applied to the training set to introduce diversity and enhance model generalization.

%\vspace{-1pt}
{FP-LGN} was evaluated against the following comparison models.
First, an ablation study was conducted based on a {C-LGN} baseline model to evaluate the impact of the feature pyramid structure in the grounding model architecture. 
In particular, {C-LGN} utilized ResNet34 \cite{he2016deep} as its contextual map feature extractor, replacing the FPN component used in {FP-LGN}. 
In addition, a human benchmark model ({Expert}) was employed by adopting the rule-based likelihood functions defined by experts in the previous work \cite{Zheng2021-zi,Fasola2013-gf,OKeefe2003,OKeefe1996}.
All tunable parameters in Expert likelihood models were optimized on the same human-annotated training dataset via maximum likelihood estimation. 
Finally, {Chance} baseline model output the likelihood randomly sampled from a uniform distribution.

\begin{table}[!t]
    \caption{NLL result summary for each comparison model.}
    \label{tab:nll_results}
    \centering
    %test \vspace{-5pt}
    \begin{tabular}{lcccc}
        \toprule
         & FP-LGN & Expert & C-LGN & Chance \\
        \midrule
        Mean & 0.384 & 0.387 & 0.532 & 1.015 \\
        SD & 0.676 & 0.881 & 0.538 & 1.012 \\
        \bottomrule
    \end{tabular}
\vspace{1pt}
\end{table}

% 7 apr 2025 version
%test \vspace{-7pt}
The statistics of the negative log-likelihood (NLL) results for each model are shown in Table~\ref{tab:nll_results}. FP-LGN achieved the lowest mean NLL of 0.384 among all comparison models, with a standard deviation of 0.676. Expert followed closely with a mean NLL of 0.387 and an SD of 0.881. C-LGN produced a higher mean NLL of 0.532 and the lowest SD of 0.538, while the Chance model produced the highest mean of 1.015 along with an SD of 1.012. 

\vspace{-1pt}
The ablation result showed that C-LGN %, omitting the feature pyramid architecture, \textbf{}
produced a significantly higher mean NLL than FP-LGN ($p < 0.01$). This result highlights the contribution of the feature pyramid architecture in improving model performance. The relatively low SD of C-LGN is attributed to repeated failures on similar input patterns, especially in cases requiring fine-grained map resolution. Next, the comparison between FP-LGN and Expert showed no significant difference in mean NLL ($p > 0.01$), 
\textcolor{black}{suggesting that the learning-based model can achieve a comparable performance in information loss to human expert in estimating the groundtruth likelihood distribution.}
Moreover, the lower SD in FP-LGN's NLL results indicates a greater robustness across diverse inputs, whereas the Expert model produced a higher \textcolor{black}{SD}, suggesting its higher sensitivity to unanticipated variations and nuances inherent in spatial language semantics. 
Finally, all tested models significantly outperformed the Chance baseline in terms of mean NLL ($p < 0.01$).

\begin{figure}[t]
%\vspace{-10pt}
%test \vspace{-2pt}
\centering
\includegraphics[width=\linewidth]{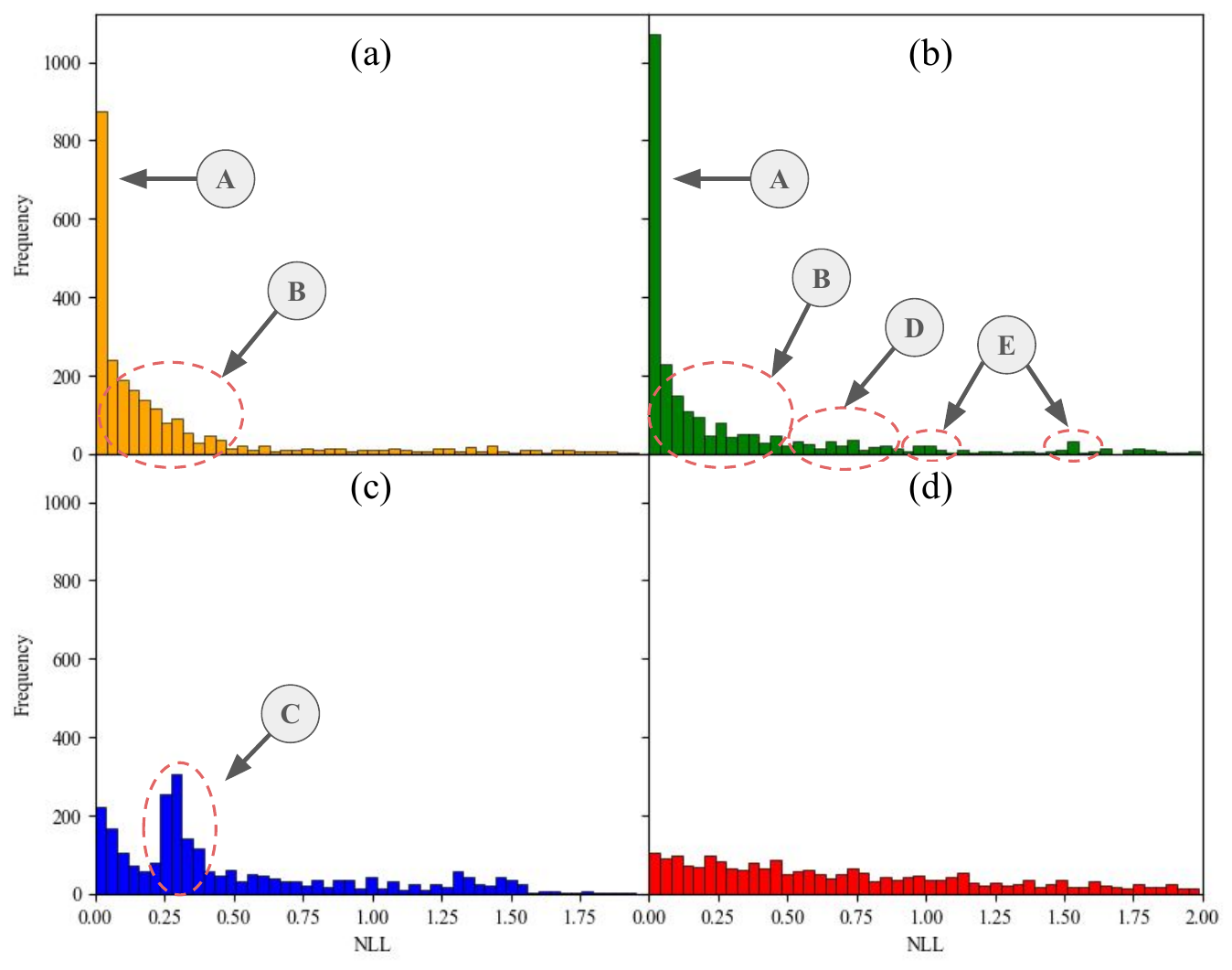}
\vspace{-20pt}
%smc\vspace{-20pt}
\caption{Histograms of NLL values in the range $[0.00, 2.00]$: (a) FP-LGN, (b) Expert, (c) C-LGN, and (d) Chance.}
\label{fig:nll_histogram}
\end{figure}

% 7 Apr 2025 version
\vspace{-1pt}
To visualize the variability in model performance, Fig.~\ref{fig:nll_histogram} shows the histograms of NLL distributions for the comparison models. 
The C-LGN histogram shows a heavier-tailed distribution and a notable secondary mode (C), indicating a degradation of probability quality compared to FP-LGN. This behavior can be attributed to the model’s limited ability to capture spatial features at different scales. In particular, precise likelihood prediction for regions located inside or near the edge of a building requires fine-grained geometric information.
By including an FPN, FP-LGN exhibited a unimodal distribution with lower dispersion histogram distribution, reflecting improved robustness and overall performance relative to learning-based models without an FPN.
Next, FP-LGN was compared against human expert. Overall, the histograms of FP-LGN and Expert both exhibited a similar shape, peaking at NLL values close to zero and decreasing in frequency toward higher NLL results, while performing similarly on average. The main difference lies in the extreme nature of the results. Specifically, Expert model histogram is relatively more prominent than FP-LGN towards the two extremes of the spectrum, i.e., in the lower NLL region where the model fitted the data points exceptionally well (A), and higher NLL region where the model failed to accommodate the data (E). 
In contrast, FP-LGN exhibited fewer predictions at the extremes, with a greater proportion of its outputs concentrated in the moderately low NLL region below 0.50 (B). As a consequence, FP-LGN shows a thinner tail than Expert (D), indicating more steady and robust performance. These results suggest that FP-LGN performs comparably to the human expert benchmark, while offering a greater performance consistency across varying semantic interpretations.

\begin{figure*}[htbp]
  \centering
  \includegraphics[width=0.89\textwidth, height=1.48in]{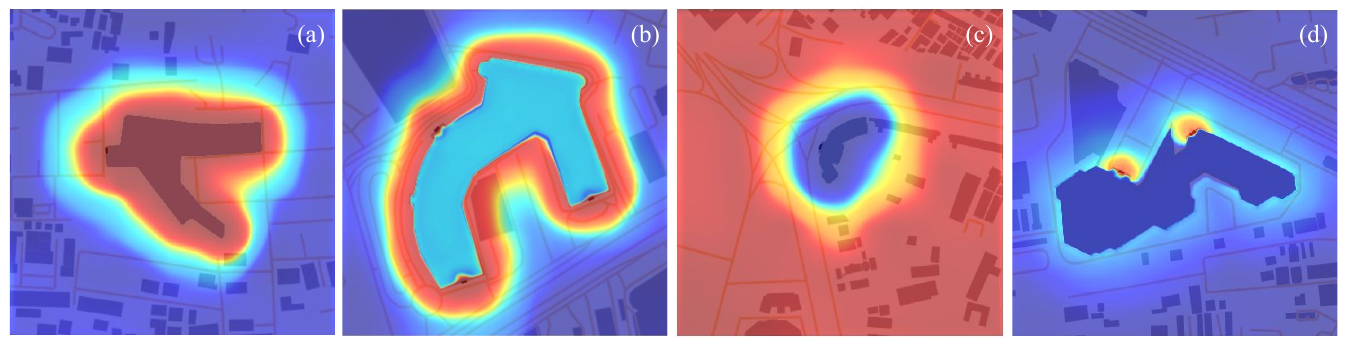}
  \vspace{-10pt}
  \caption{Example likelihood grounding learned by FP-LGN for spatial relations (a) ``at," (b) ``near," (c) ``far from," and (d) ``in front of."}% across maps
  \label{fig:result_all}
  \vspace{-18pt}
\end{figure*}

% \vspace{-10pt}
\textcolor{black}{Fig.~\ref{fig:result_all} shows examples of the learned FP-LGN likelihood outputs. It was found that FP-LGN successfully grounded the likelihood, capturing spatial semantics according to the geometric properties of reference landmarks. This is reflected in distinct relationships displayed between the output likelihood distributions and the reference landmark's geometrical structure.} 
For instance, the semantics likelihood of spatial relations such as ``at," ``near," and ``far from" generally followed the shapes of the reference landmark's contours. Some examples of these are shown in \ref{fig:result_all}(a)-(b). 
However, an interesting behavior emerged in the cases where the reference landmarks were small concave buildings such as in Fig. \ref{fig:result_all}(c). It was found that the FP-LGN likelihood learned from data followed the convex hulls of the landmarks instead of their concave silhouettes. 
This pattern was found to match the behavior in the spatial semantics data provided by humans. 
In contrast, this adaptive behavior was not observed in the likelihood outputs of Expert model which was found to consistently produce concave-shaped likelihood distributions, simply scaling them with the building sizes.
This finding demonstrates the advantage of the learning-based approach that is data-driven, allowing flexibility through adjustment to complex and nuanced patterns in human semantics.

% \vspace{-10pt}
\textcolor{black}{In summary, FP-LGN was found to successfully ground spatial language likelihood by learning directly from human data, achieving an information loss comparable to the likelihood model manually defined by human experts as indicated by no significant difference in mean NLL, as well as demonstrating a greater robustness in grounding performance indicated by a lower NLL standard deviation. The proposed FPN feature extractor component enables multiresolution feature extraction of environment map, improving upon the ablation baseline.}

\vspace{-4.5pt}
\subsection{Human-Robot Collaborative Sensing: Target Search Tasks}
\vspace{-2.5pt}
\begin{figure}[t]
\centering
\includegraphics[width=\linewidth]{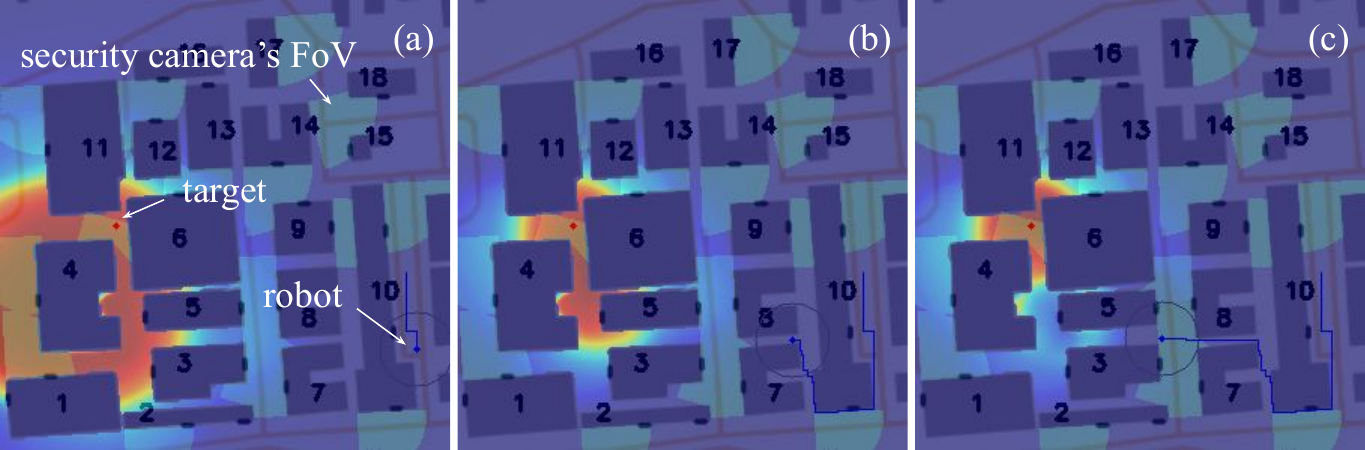}
\vspace{-20pt}
\caption{The human interface showing the evolving target posterior belief after recursive Bayesian updates given a cumulative sequence of multiple spatial language observations in the following order: (a) “you can find the bag around building 4,” (b) “the bag's close to building 6,” and (c) “the bag's not in front of building 5.” Given the input sequence, the posterior converged toward the true location of the target.}
\label{fig:fusion_sequence_example}
\end{figure}

To evaluate task performances when leveraging the spatial language likelihood grounding for human-robot collaborative information gathering,
a motivating simulated target search scenario setup inspired by the previous work
\cite{TseTRO2018,Burks2021-im} was followed.
In this scenario, a mixed human-robot security team was tasked to search for a hidden target, i.e., a reported suspicious bag.  
The goal of the human-robot team was to locate the hidden target as quickly as possible. 
An autonomous mobile robot was deployed to the search region to gather information on the target by actively sensing the environment via its onboard camera. 
Simultaneously, a human security personnel was able to monitor the search environment remotely via surveillance cameras and communicated their observations to the robot in natural language sentences. 
Through collaborative sensing, 
the robot recursively fused its sensor measurements with human spatial language observations using the Bayesian update equations~\eqref{eqn_robot_measurement_update_step}-\eqref{eqn_human_measurement_update_step}.
The robot’s decision making was performed according to the fused posterior distribution representing the target's estimate given all information aggregated over time from all sources.
The search was successful when the robot captured the target within its camera’s field of view. 
One hundred and fifty search scenarios were performed using the OSM maps extracted from three cities in Thailand, with each search initialized using randomized robot and target positions.
Screen captures of the human interface displaying a region of search environment overlaid with the target posterior distributions are shown in \figurename~\ref{fig:fusion_sequence_example}.
Each building was named as ``Building $<$ID$>$,'' where $<$ID$>$ was the number marked on the interface. 
The human provided inputs in natural language, making either positive (e.g., “The bag is in front of Building 8.”) or negative observations (e.g., “The bag is not in front of Building 8.”). 
These inputs were parsed using LLaMA 2 7B into 
spatial observation tuples \((T_i, R_{i,k}, \gamma_{i,k})\) 
which were then used as inputs to the likelihood grounding model as described in Sec.~\ref{training_parsing}.
The security cameras were positioned around the map with a $45 ^\circ$ fixed cone Field of View (FoV) giving a partial view of the environment. The target's position was revealed on the map to the human only if the target was in the FoV of any security camera. 
The robot ran at a speed of $1\ \mathrm{m}/\mathrm{s}$, while 
the target detector ran on the $360^\circ$ camera inputs at $1\,\mathrm{Hz}$ with the true positive and true negative rates of $0.8$ within the detection range of $25\,\mathrm{m}$. 
In each time step, the robot planned its path towards the current Maximum A Posterior (MAP) estimate of the target position using the A* algorithm.
Search performance was evaluated by the percentage of successful searches within a limited number of search steps.
Four information gathering modes were conducted: robot-only (no human inputs; robot only fused its own sensor measurements), human-only (no sensor inputs; robot only fused human language inputs), collaborative human-robot (both sensor and human inputs were fused), and uninformed (neither human observations nor sensor measurements were fused).

To contextualize the results, the human input sentences are briefly summarized as follows. 
First, it was found that most sentences (83.0\%) followed a subject-predicate structure, with the subject being either the target (e.g., ``The bag is near Building 7.'') or the robot (e.g., ``You can find the bag near Building 7.''). The rest (17.0\%) followed an existential structure (e.g., ``There is a bag in front of Building 16.'').
The sentences contained a variety of spatial prepositions, most commonly ``in front of,'' ``near,'' ``close to," ``beside,'' ``next to,'' ``around,'' and ``alongside.'' These were often modified by negation (e.g., ``not,'' “nowhere''). 
Most verbs were in active voice (69.7\%), with the remaining in passive voice (e.g., ``can be found"). 
The verb ``is'' was shortened to ``'s'' in about half of its appearances. 
Additionally, 32.5\% of sentences contained at least one typographical error. 
The parser was able to parse most sentences correctly, achieving an accuracy rate of 97.7\%.
\textcolor{black}{The parsed observation tuple was then passed on to FP-LGN for likelihood grounding and fusion processes. The results of the collaborative target search are discussed next.}

\begin{figure}[t]
\vspace{-3pt}
\centering
\includegraphics[width=2.23in]{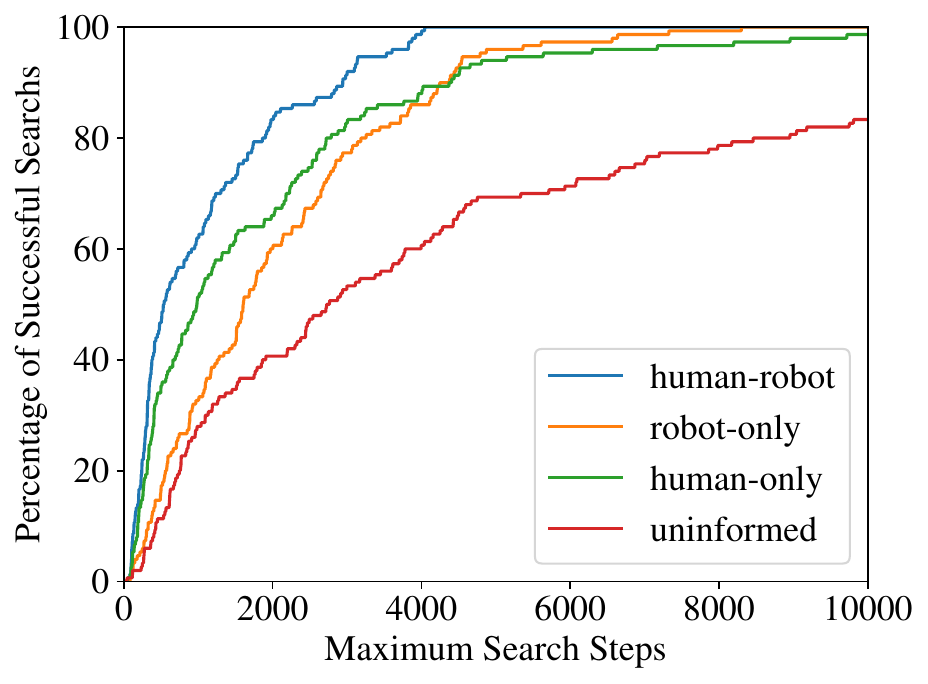}
%\includegraphics[width=2.23in]{search_curve_4may2025.pdf}
%\includegraphics[width=2.23in, height=1.65in]{search_curve_4may2025.pdf}
%\includegraphics[width=2.23in, height=1.60in]{search_curve_4may2025.pdf}
%SMC\vspace{-12pt}
\vspace{-14pt}
\caption{Comparison of successful search percentages among the four types of input information modes.}
\label{fig:seach_performance_curve}
\end{figure}

% \vspace{-1pt}
\figurename~\ref{fig:seach_performance_curve} presents the search performance results, showing the percentage of successful searches versus maximum search steps. First, the human-robot collaborative information gathering mode was able to achieve a 100.0\% success rate within 4043 steps. In comparison, the robot-only mode required 105.0\% longer search limit of 8300 steps to reach 100.0\% success rate. On the other hand, human-only and uninformed modes failed to reach 100.0\% success rate within 10,000 search steps limit. It was found that the percentage of successful searches for the human-only mode plateaued at 94.0\% after 4807 steps. 
Similarly, robot-only performances plateaued at 96.0\% after 4872 steps. 
In contrast, the collaborative mode, integrating human language observations and robot sensor measurements, was able to overcome these limitations.

\vspace{-0.5pt}
These results demonstrated that the robot successfully fused information in the heterogeneous forms of human language observations and sensor measurements, leveraging collaborative sensing benefits through the complementary perceptual capabilities of the human and the robot.
Noticeably, this was achieved as fusing robot sensor measurements helped reduce uncertainty in search regions where human language observation lacked specificity. 
Additionally, it was observed that human-robot mode achieved greater success rate than human-only and robot-only modes at all search step limits.  
In missions requiring fewer steps, the human-only mode outperformed the robot-only mode and approached the performance of the human-robot mode. This suggests that when the target location could be clearly described, human inputs were highly effective in helping the robot improve its task performance. 

\vspace{-0.5pt}
%\vspace{100pt}
The benefit of human collaboration also increased when multiple spatial language inputs were recursively fused, as illustrated by the target posterior distribution in \figurename~\ref{fig:fusion_sequence_example}.
Each spatial language observation by human provided additional information, contributing to the decrease in the target estimate's uncertainty, as reflected in a more tightly concentrated, i.e., lower entropy, target posterior. 
\textcolor{black}{The recursive reduction in the target estimate’s uncertainty according to the input language semantics demonstrated the robot’s ability to perform probabilistic reasoning over multiple spatial language sentences. } 
\textcolor{black}{This probabilistic Bayesian reasoning approach provided interpretable means for human users to understand how the robot’s target estimate evolved over time.}

\vspace{-0.5pt}
%\vspace{-3pt}
Finally, to summarize the overall performance of each information gathering mode, the mean and standard deviation of the search steps are shown in Table~\ref{tab:search_step_results}. 
The mean number of search steps taken to complete the task 
when collaborative sensing was performed between human and robot was 1054 steps, which reduced by 47.8\%, 41.6\%, and 87.4\% compared to the robot-only (2021 steps), human-only (1804 steps), and uninformed modes (8411 steps), respectively. All of these reductions were found to be statistically significant ($p < 0.01$). 
Finally, it was observed that fusing information from both sources allowed greater robustness to task variations, as indicated by the lower standard deviation (1065 steps) compared to those using the other information gathering modes (1610, 2273, and 18739 steps).

\textcolor{black}{In conclusion, the results demonstrated that the grounded likelihood successfully enabled uncertainty-aware fusion of heterogeneous human language observations and sensor measurements, yielding significant improvements in human-robot collaborative sensing task performance. 
Furthermore, the recursive Bayesian fusion approach allows an interpretable probabilistic reasoning that refines the target posterior over multiple observations, thereby reducing the output estimation uncertainty over time.}

\begin{table}[!t]
\vspace{-5pt}
    \caption{Number of search steps for each information gathering mode.}
    \label{tab:search_step_results}
    \centering
    \vspace{-5pt}
    \begin{tabular}{lcccc}
        \toprule
         & human-robot & robot-only & human-only & uninformed \\
        \midrule
        Mean & 1054 & 2021 & 1804 & 8411 \\
        SD & 1065 & 1610 & 2273 & 18739 \\
        \bottomrule
    \end{tabular}
\end{table}
%\vspace{-6 pt}
\vspace{-5 pt}
%\vspace{-6 pt}
\section{Conclusions}
\vspace{-4 pt}
This paper proposes the first fully-learnable spatial language grounding model for grounding spatial language likelihood from contextual map inputs, by learning the geometric map features and their relationship to spatial relation semantics, allowing the likelihood to automatically adapt to unseen landmarks.
Trained as a probability estimator, the model captures the aleatoric uncertainty in human language, achieving information loss performance comparable to a likelihood grounding model written by human experts, while exhibiting greater robustness.
\textcolor{black}{Additionally, results showed that the learned likelihood was successfully used for uncertainty-aware fusion of human language observations and robot sensor measurements, achieving significant improvements in human–robot collaborative sensing task performance. 
The recursive Bayesian fusion approach allowed interpretable probabilistic reasoning that refined the target posterior distribution over multiple observations, thereby reducing its uncertainty over time.}
\vspace{-2 pt}

%\vspace{-1pt}
%\vspace{-3pt}
%SMC \vspace{-3pt}
\bibliographystyle{IEEEtran}
\vspace{-4pt}
\bibliography{refs}

% Generated by IEEEtran.bst, version: 1.14 (2015/08/26)
\begin{thebibliography}{10}
\providecommand{\url}[1]{#1}
\csname url@samestyle\endcsname
\providecommand{\newblock}{\relax}
\providecommand{\bibinfo}[2]{#2}
\providecommand{\BIBentrySTDinterwordspacing}{\spaceskip=0pt\relax}
\providecommand{\BIBentryALTinterwordstretchfactor}{4}
\providecommand{\BIBentryALTinterwordspacing}{\spaceskip=\fontdimen2\font plus
\BIBentryALTinterwordstretchfactor\fontdimen3\font minus
  \fontdimen4\font\relax}
\providecommand{\BIBforeignlanguage}[2]{{%
\expandafter\ifx\csname l@#1\endcsname\relax
\typeout{** WARNING: IEEEtran.bst: No hyphenation pattern has been}%
\typeout{** loaded for the language `#1'. Using the pattern for}%
\typeout{** the default language instead.}%
\else
\language=\csname l@#1\endcsname
\fi
#2}}
\providecommand{\BIBdecl}{\relax}
\BIBdecl

\bibitem{probRoboticsBook}
S.~Thrun, W.~Burgard, and D.~Fox, \emph{Probabilistic robotics}.\hskip 1em plus
  0.5em minus 0.4em\relax {MIT} Press, 2005.

\bibitem{BarShalom}
Y.~Bar-Shalom, X.~R. Li, and T.~Kirubarajan, \emph{Estimation with Applications
  to Tracking and Navigation}.\hskip 1em plus 0.5em minus 0.4em\relax John
  Wiley \& Sons, 2001.

\bibitem{TseRAL19_short}
Y.~Han, R.~Tse, and M.~Campbell, ``Pedestrian motion model using non-parametric
  trajectory clustering and discrete transition points,'' \emph{{IEEE} Robot.
  Autom. Lett.}, vol.~4, no.~3, pp. 2614--2621, 2019.

\bibitem{fusion_PF_AUV_Tracking_RAL24}
S.~Chun, C.~Kawamura, K.~Ohkuma, and T.~Maki, ``3{D} detection and tracking of
  a moving object by an autonomous underwater vehicle with a multibeam imaging
  sonar: Toward continuous observation of marine life,'' \emph{IEEE Rob. Autom.
  Lett.}, vol.~9, no.~4, pp. 3037--3044, 2024.

\bibitem{fusion_ESKF_CuRobot_localize_RAL24}
K.~Yang~et al., ``Design and trajectory tracking control of {C}u{R}obot: A
  cubic reversible robot,'' \emph{IEEE Rob. Autom. Lett.}, vol.~9, no.~4, pp.
  3029--3036, 2024.

\bibitem{Wyffels2015-ul}
K.~Wyffels and M.~Campbell, ``Negative information for occlusion reasoning in
  dynamic extended multiobject tracking,'' \emph{IEEE Trans. Robot.}, vol.~31,
  no.~2, pp. 425--442, Apr. 2015.

\bibitem{Wakayama2023}
S.~Wakayama and N.~Ahmed, ``Probabilistic semantic data association for
  collaborative human-robot sensing,'' \emph{IEEE Trans. Robot.}, 2023.

\bibitem{EKF_IMU_Odom_2DLiDAR_IROS_2020}
P.~Geneva, N.~Merrill, Y.~Yang, C.~Chen, W.~Lee, and G.~Huang, ``Versatile 3{D}
  multi-sensor fusion for lightweight 2{D} localization,'' in \emph{Proc. IEEE
  Int. Conf. Intell. Robots Syst.}, 2020, pp. 4513--4520.

\bibitem{Charrow2014-tq}
B.~Charrow, N.~Michael, and V.~Kumar, ``Cooperative multi-robot estimation and
  control for radio source localization,'' \emph{Int. J. Rob. Res.}, vol.~33,
  no.~4, pp. 569--580, Apr. 2014.

\bibitem{KF_singleCam_MOT_DataAssoc_IROS22}
R.~Ge, M.~Lee, V.~Radhakrishnan, Y.~Zhou, G.~Li, and G.~Loianno, ``Vision-based
  relative detection and tracking for teams of micro aerial vehicles,'' in
  \emph{Proc. Int. Conf. Intell. Robots Syst.}, 2022, pp. 380--387.

\bibitem{EKFvisualTrackingICRA24}
A.~Saviolo, P.~Rao, V.~Radhakrishnan, J.~Xiao, and G.~Loianno, ``Unifying
  foundation models with quadrotor control for visual tracking beyond object
  categories,'' in \emph{Proc. {IEEE} Int. Conf. Robot. Autom.}, 2024.

\bibitem{ferrari23fusion}
K.~A. LeGrand and S.~Ferrari, ``Split happens! imprecise and negative
  information in {G}aussian mixture random finite set filtering.'' \emph{J.
  Adv. Inf. Fusion}, vol.~17, no.~2, 2022.

\bibitem{TseTRO2015}
R.~Tse, N.~R. Ahmed, and M.~Campbell, ``Unified terrain mapping model with
  {M}arkov {R}andom {F}ields,'' \emph{IEEE Trans. Robot.}, vol.~31, no.~2, pp.
  290--306, 2015.

\bibitem{TseMFI2012}
R.~Tse, N.~Ahmed, and M.~Campbell, ``Unified mixture-model based terrain
  estimation with {M}arkov {R}andom {F}ields,'' in \emph{Proc. {IEEE} Int.
  Conf. Multisensor Fusion Integr. Intell. Syst.}, 2012, pp. 238--243.

\bibitem{Ahmed2013-mm}
N.~Ahmed, E.~M. Sample, and M.~Campbell, ``{B}ayesian multicategorical soft
  data fusion for {Human--Robot} collaboration,'' \emph{IEEE Trans. Robot.},
  vol.~29, no.~1, pp. 189--206, Feb. 2013.

\bibitem{NisarMMS_ACC08}
N.~Ahmed and M.~Campbell, ``Multimodal operator decision models,'' in
  \emph{Proc. Amer. Control Conf.}, 2008, pp. 4504--4509.

\bibitem{NisarICRA2010}
------, ``Variational {B}ayesian data fusion of multi-category discrete
  observations, with applications to cooperative human-robot estimation,'' in
  \emph{Proc. {IEEE} Int. Conf. Robot. Autom.}, 2010.

\bibitem{AhmedAAAI14}
N.~R. Ahmed, R.~Tse, and M.~Campbell, ``Enabling robust human-robot cooperation
  through flexible fully {B}ayesian shared sensing,'' in \emph{AAAI Spring
  Symposium Series}, 2014.

\bibitem{TseTRO2018}
R.~Tse and M.~Campbell, ``Human–robot communications of probabilistic beliefs
  via a {D}irichlet process mixture of statements,'' \emph{IEEE Trans. Robot.},
  vol.~34, no.~5, pp. 1280--1298, 2018.

\bibitem{TseIROS2015}
------, ``Human-robot information sharing with structured language generation
  from probabilistic beliefs,'' in \emph{Proc. IEEE Int. Conf. Intell. Robots
  Syst.}, 2015, pp. 1242--1248.

\bibitem{NisarExpertSyst2012}
N.~Ahmed and M.~Campbell, ``On estimating simple probabilistic discriminative
  subclass models,'' \emph{Expert Syst. Appl.}, vol.~39, 2012.

\bibitem{BishopLanguage2013}
A.~N. Bishop and B.~Ristic, ``Fusion of spatially referring natural language
  statements with random set theoretic likelihoods,'' \emph{{IEEE} Trans.
  Aerosp. Electron. Syst.}, vol.~49, no.~2, pp. 932--944, 2013.

\bibitem{BishopFusion2011}
------, ``Fusion of natural language propositions: {B}ayesian random set
  framework,'' in \emph{Proc. Int. Conf. Inf. Fusion}, 2011, pp. 1--8.

\bibitem{BishopRistic2011}
------, ``Spatially referring natural language propositions: Information fusion
  and estimation theory,'' in \emph{Proc. U.S./Australia Joint Workshop Def.
  Appl. Signal Process. ({DASP})}, 2011, pp. 1--11.

\bibitem{arulampalam18spatial_natural_language_binary_model}
S.~Arulampalam, B.~Ristic, and J.~Legg, ``Learning the parameters of
  spatially-referring natural language likelihoods in binary models,'' in
  \emph{Proc. Int. Conf. Inf. Fusion}, 2018, pp. 1011--1017.

\bibitem{OpenStreetMap}
\BIBentryALTinterwordspacing
{OpenStreetMap contributors}, ``{OpenStreetMap},'' 2024. [Online]. Available:
  \url{https://www.openstreetmap.org}
\BIBentrySTDinterwordspacing

\bibitem{Sweet2016-jm}
N.~Sweet and N.~Ahmed, ``Structured synthesis and compression of semantic human
  sensor models for {B}ayesian estimation,'' in \emph{{Proc. Amer. Control
  Conf.}}, 2016, pp. 5479--5485.

\bibitem{Ahmed2018-mw}
N.~Ahmed, ``{Data-Free/Data-Sparse} softmax parameter estimation with
  structured class geometries,'' \emph{IEEE Signal Process. Lett.}, vol.~25,
  no.~9, pp. 1408--1412, Sep. 2018.

\bibitem{Burks2023}
L.~Burks, H.~M. Ray, J.~McGinley, S.~Vunnam, and N.~Ahmed, ``Harps: An online
  pomdp framework for human-assisted robotic planning and sensing,'' \emph{IEEE
  Trans. Robot.}, vol.~39, no.~4, pp. 3024--3042, 2023.

\bibitem{Newman2009}
J.~Frost, ``Mapping spatial language to sensor models,'' in \emph{Comput. Lab.
  Student Conf.}, 2009.

\bibitem{Newman2010}
J.~Frost, A.~Harrison, S.~Pulman, and P.~Newman, ``A probabilistic approach to
  modelling spatial language with its application to sensor models,'' in
  \emph{Proc. {COSLI} Workshop}, 2010.

\bibitem{Zheng2021-zi}
K.~Zheng, D.~Bayazit, R.~Mathew, E.~Pavlick, and S.~Tellex, ``Spatial language
  understanding for object search in partially observed city-scale
  environments,'' in \emph{Proc. IEEE Int. Symp. Robot Human Interact.
  Commun.}, 2021, pp. 315--322.

\bibitem{OKeefe1996}
J.~O'Keefe and N.~Burgess, ``Geometric determinants of the place fields
  hippocampal neurons,'' \emph{Nature}, vol. 381, pp. 425--8, Jun. 1996.

\bibitem{OKeefe2003}
J.~O'Keefe, \emph{{Vector Grammar, Places, and the Functional Role of the
  Spatial Prepositions in English}}.\hskip 1em plus 0.5em minus 0.4em\relax
  Oxford Univ. Press, 2003.

\bibitem{Fasola2013-gf}
J.~Fasola and M.~J. Mataric, ``Using semantic fields to model dynamic spatial
  relations in a robot architecture for natural language instruction of service
  robots,'' in \emph{Proc. IEEE Int. Conf. Intell. Robots Syst.}, 2013.

\bibitem{pmlr-v162-liu22f}
S.~Liu~et al., ``Deep probability estimation,'' in \emph{Proc. Int. Conf. Mach.
  Learn.}, vol. 162, 2022, pp. 13\,746--13\,781.

\bibitem{Lin2016FeaturePN}
T.-Y. Lin, P.~Doll{\'a}r, R.~B. Girshick, K.~He, B.~Hariharan, and S.~J.
  Belongie, ``Feature pyramid networks for object detection,'' \emph{Proc.
  IEEE/CVF Conf. Comput. Vis. Pattern Recognit.}, pp. 936--944, 2017.

\bibitem{ren2016faster}
S.~Ren, K.~He, R.~Girshick, and J.~Sun, ``Faster {R-CNN}: Towards real-time
  object detection with region proposal networks,'' \emph{IEEE Trans. Pattern
  Anal. Mach. Intell.}, vol.~39, no.~6, pp. 1137--1149, 2016.

\bibitem{blukis2019learning}
V.~Blukis, Y.~Terme, E.~Niklasson, R.~A. Knepper, and Y.~Artzi, ``Learning to
  map natural language instructions to physical quadcopter control using
  simulated flight,'' in \emph{Proc. Conf. Robot Learn. (CoRL)}, ser. Proc.
  Mach. Learn. Res., vol. 100, 2020, pp. 1415--1438.

\bibitem{tellex2011understanding}
S.~Tellex, T.~Kollar, S.~Dickerson, M.~Walter, A.~Banerjee, S.~Teller, and
  N.~Roy, ``Understanding natural language commands for robotic navigation and
  mobile manipulation,'' in \emph{Proc. AAAI Conf. Artif. Intell.}, vol.~25,
  no.~1, 2011, pp. 1507--1514.

\bibitem{brown2020language}
T.~Brown~et al., ``Language models are few-shot learners,'' in \emph{Adv.
  Neural Inf. Process. Syst. ({NeurIPS})}, vol.~33, 2020, pp. 1877--1901.

\bibitem{zakharov2019few}
E.~Zakharov, A.~Shysheya, E.~Burkov, and V.~Lempitsky, ``Few-shot adversarial
  learning of realistic neural talking head models,'' in \emph{Proc. IEEE/CVF
  Int. Conf. Comput. Vis.}, 2019, pp. 9459--9468.

\bibitem{li2023revisiting}
G.~Li, P.~Wang, and W.~Ke, ``Revisiting large language models as zero-shot
  relation extractors,'' in \emph{Proc. Conf. Empir. Methods NLP}, 2023.

\bibitem{touvron2023llama}
H.~Touvron~et al., ``Llama 2: Open foundation and fine-tuned chat models,''
  \emph{arXiv preprint arXiv:2307.09288}, 2023.

\bibitem{prolific_2025}
\BIBentryALTinterwordspacing
{Prolific}, ``{Prolific} participant recruitment platform,'' 2025. [Online].
  Available: \url{https://www.prolific.com}
\BIBentrySTDinterwordspacing

\bibitem{he2016deep}
K.~He, X.~Zhang, S.~Ren, and J.~Sun, ``Deep residual learning for image
  recognition,'' in \emph{Proc. IEEE Conf. Comput. Vis. Pattern Recognit.},
  2016.

\bibitem{Burks2021-im}
L.~Burks~et al., ``Collaborative human-autonomy semantic sensing through
  structured {POMDP} planning,'' \emph{Rob. Auton. Syst.}, vol. 140, p. 103753,
  Jun. 2021.

\end{thebibliography}
\vspace{-12pt}
% \vspace{12pt}

\end{document}